\documentclass[letterpaper]{article} % DO NOT CHANGE THIS
\usepackage{aaai2026}  % DO NOT CHANGE THIS  [submission]
\usepackage{times}  % DO NOT CHANGE THIS
\usepackage{helvet}  % DO NOT CHANGE THIS
\usepackage{courier}  % DO NOT CHANGE THIS
\usepackage[hyphens]{url}  % DO NOT CHANGE THIS
\usepackage{graphicx} % DO NOT CHANGE THIS
\usepackage{natbib}  % DO NOT CHANGE THIS AND DO NOT ADD ANY OPTIONS TO IT
\usepackage{caption} % DO NOT CHANGE THIS AND DO NOT ADD ANY OPTIONS TO IT
\usepackage{algorithm}
\usepackage{algorithmic}
\usepackage{amsmath}
\usepackage{booktabs} 
\usepackage{multirow}
\usepackage{array}
\usepackage{soul, color, xcolor}
\usepackage{threeparttable}
\usepackage{newfloat}
\usepackage{listings}
\DeclareCaptionStyle{ruled}{labelfont=normalfont,labelsep=colon,strut=off} % DO NOT CHANGE THIS
\floatstyle{ruled}
\newfloat{listing}{tb}{lst}{}
\floatname{listing}{Listing}
\title{SDSNN: A Single-Timestep Spiking Neural Network with Self-Dropping Neuron and Bayesian Optimization}
\author{
    Changqing Xu, Buxuan Song, Yi Liu, Xinfang Liao, Wenbin Zheng, Yintang Yang
}
\affiliations{
    School of Microelectronics, Xidian University, Xi’an, China \\
    \{cqxu, bxsong, yiliu, xfliao, wbzheng, ytyang\}@xidian.edu.cn
}

\usepackage{bibentry}
\begin{document}

\maketitle

\begin{abstract}
Spiking Neural Networks (SNNs), as an emerging biologically inspired computational model, demonstrate significant energy efficiency advantages due to their event-driven information processing mechanism. Compared to traditional Artificial Neural Networks (ANNs), SNNs transmit information through discrete spike signals, which substantially reduces computational energy consumption through their sparse encoding approach. However, the multi-timestep computation model significantly increases inference latency and energy, limiting the applicability of SNNs in edge computing scenarios. We propose a single-timestep SNN, which enhances accuracy and reduces computational energy consumption in a single timestep by optimizing spike generation and temporal parameters. We design a Self-Dropping Neuron mechanism, which enhances information-carrying capacity through dynamic threshold adjustment and selective spike suppression. Furthermore, we employ Bayesian optimization to globally search for time parameters and obtain an efficient inference mode with a single time step. Experimental results on the Fashion-MNIST, CIFAR-10, and CIFAR-100 datasets demonstrate that, compared to traditional multi-timestep SNNs employing the Leaky Integrate-and-Fire (LIF) model, our method achieves classification accuracies of 93.72\%, 92.20\%, and 69.45\%, respectively, using only single-timestep spikes, while maintaining comparable or even superior accuracy. Additionally, it reduces energy consumption by 56\%, 21\%, and 22\%, respectively.
\end{abstract}

\section{Introduction}
%1.脉冲神经网络的生物学合理性与低功耗优势
%2.传统LIF神经元模型脉冲触发机制的局限性：脉冲频率骤增，形成冗杂脉冲序列
%3.时间步长是影响SNN信息处理能力的关键因素，SNN通常需要多个时间步来训练和推理，导致延迟增加和计算开销
%4.时间步长调优挑战：时间步长固定或启发式选择的低效性
%5.我们工作的怎么做，及创新点（
In recent years, Spiking Neural Networks (SNNs) have attracted widespread attention due to their biological compatibility and potential for ultra-low power consumption. SNNs transmit information through spikes instead of continuous signals, exhibiting significant computational sparsity, which provides them with unique advantages in terms of energy efficiency. Among existing SNNs, the Leaky Integrate-and-Fire (LIF) model is the most commonly used spiking neuron. Various variants with different biological characteristics have been developed to address challenges encountered by SNNs in different tasks \cite{yao2022glif, huang2024clif, lian2024lif, zhang2024tc}, along with new neuron models proposed from a biological perspective \cite{yang2022sam, wang2022signed, li2022efficient} that aim to solve challenges related to training efficiency, performance, and biological plausibility in SNNs. However, many of these works increase model complexity.

Currently, the mainstream methods for training SNNs include gradient-based surrogate methods and ANN-to-SNN conversion \cite{cao2015spiking, bu2023optimal, meng2022training,deng2021optimal}. The gradient-based surrogate method approximates the non-differentiable spike function by introducing a surrogate gradient, enabling gradient-based optimization. However, this method requires Backpropagation Through Time (BPTT) during training, resulting in gradient updates across multiple time steps, significantly increasing computational overhead. In contrast, the ANN-to-SNN conversion method directly maps the pre-trained ANN parameters into the SNN. While this reduces training time to some extent, it requires more inference time steps to achieve higher accuracy, inevitably increasing inference latency. Therefore, reducing computation and inference latency while maintaining performance has become a key challenge in SNN research.

The energy efficiency of SNNs primarily stems from the computational sparsity, where addition operations replace multiplication and accumulation (MAC) operations in ANNs. However, unlike ANNs, which complete inference in a single time step, SNNs require multiple time steps for stepwise computation, leading to higher inference latency. Currently, the time steps in SNNs are typically fixed or heuristically selected as hyperparameters before training begins \cite{hu2024advancing, xu2023constructing, meng2022training,su2023deep}. The number of time steps as a hyperparameter has a significant impact on network performance. Different time steps contain different temporal information. Too few time steps may result in performance degradation, while too many time steps lead to redundant computations, increasing unnecessary consumption.

In this paper, we propose a single-timestep SNN to reduce the latency of SNNs. We design a Self-Dropping Neuron model, which generates spikes when the membrane potential exceeds the threshold and starts to drop, aiming to increase the information carried by individual spikes. Additionally, we only compute gradients on the time steps for backpropagation, eliminating the computational complexity of gradients over the temporal domain. By combining Bayesian optimization, we consider the heterogeneity between spike layers and automatically search for the optimal time step combinations for each layer. Based on this, we propose a three-stage optimization method that enables a more precise search, faster convergence, and higher accuracy through a staged optimization strategy.
The following summarizes our main contributions:

\begin{enumerate}
    \item \textbf{Self-Dropping Neuron with Spike Suppression:} We propose a Self-Dropping Neuron, which improves the information-carrying capacity of neural spikes. Since gradients are only propagated over a single time step, this approach avoids the repeated temporal gradient computations inherent in traditional multi-time-step methods, thereby further reducing computational overhead.
    
    \item \textbf{Automatic Time Parameter Tuning Using Bayesian Optimization:} We design an automatic time parameter tuning method based on Bayesian optimization. Based on single-timestep spiking neural networks, we employ Bayesian optimization to automatically search for the optimal time step combination for each spike layer, thereby achieving faster convergence and higher accuracy, while ensuring collaborative optimization between different layers of the network.
    
    \item \textbf{Evaluation on Benchmark Datasets:} We evaluate the proposed method on datasets such as Fashion-MNIST, CIFAR-10, and CIFAR-100. This method achieves comparable or even better model accuracy than traditional methods while utilizing only a single time step.
\end{enumerate}
%以上内容均需要相关文献的佐证（参考文献注意时效性和权威性）

\section{Related work}
%SNN总体的发展，还存在的不足（不足可以从我们要做的工作的角度说）
%针对与降低脉冲发放的相关工作
%其中以第三人称的方式把我们的之前的工作也介绍一下
\subsection{Spiking Neuron Models}
The basic computational unit of an SNN is the spiking neuron. The Leaky Integrate-and-Fire (LIF) model \cite{teeter2018generalized} is the most commonly used spiking neuron model in SNNs. The mathematical expression of the LIF neuron is given by 
\begin{equation}\label{eq:1}
u^{(t,n+1)} = \tau u^{(t-1,n+1)} + x^{(t,n)}
\end{equation}
where $u^{(t,n+1)}$ represents the membrane potential of the n+1 layer neuron at time t, $x^{(t,n)}$ is the input to the $n+1$ layer neuron at time t, and $\tau$ is the membrane time constant. When the membrane potential $u^{(t,n+1)}$ exceeds the threshold $Vth$, the neuron generates a spike $s[t]$ through the Heaviside function, as shown as 
\begin{equation}\label{eq:2}
u^{(t,n+1)} = \tau u^{(t-1,n+1)} + x^{(t,n)}
\end{equation}
After the spike, the neuron resets its membrane potential. Two common reset methods are shown as 
\begin{equation}\label{eq:3}
{{u}^{t,n+1}}=\{_{{{u}^{t,n+1}}-{{V}_{th}}\cdot {{O}^{t,n+1}}}^{{{u}^{t,n+1}}\cdot (1-{{O}^{t,n+1}})}
 \end{equation}
The first method is referred to as a hard reset, where the membrane potential at the spike location is reset to zero. In this paper, we adopt the second method, known as a soft reset, where the membrane potential at the spike location is reduced by a threshold voltage, allowing the neuron to retain more temporal information \cite{han2020rmp}.

\cite{zhong2022spike} introduced a lateral inhibition spiking neural network model based on Spike-Timing-Dependent Plasticity (STDP), which improves model generalization by using adaptive threshold suppression to prevent overfitting. \cite{xu2020boosting} proposed a temporal compression technique for hardware spiking neural network accelerators, where multiple adjacent binary spikes are merged into a single weighted spike. This technique, combined with the input-output-weighted spiking neuron model, facilitates the processing of weighted spikes while significantly reducing the number of spikes that need to be handled, thereby preserving spike information. \cite{liu2022spikeconverter} introduced an ANN-to-SNN conversion framework called SpikeConverter, which improves SNN inference efficiency and energy efficiency through inverse-leaky integrate-and-fire neurons, temporal separation, optimized spike encoding, and layer-wise pipelining. \cite{wu2019direct} proposed a direct training method for SNNs by narrowing the rate coding window to reduce simulation length and neuron normalization to balance the firing rates across the network. \cite{xu2022direct} introduced a direct training method for SNNs based on a multi-threshold LIF model and backpropagation (BP) algorithm, which increases the information capacity of each spike by incorporating a multi-threshold LIF model, significantly reducing the latency in SNNs.

These works mainly focus on improving model performance, generalization, and inference efficiency. However, they lack consideration of the neuronal perspective, which increases the complexity and implementation cost of the design.

\subsection{Time Steps of Spiking Neural Networks}
Recent works have optimized SNN inference energy consumption by employing various techniques to reduce the number of time steps. The work in \cite{chowdhury2021one} begins by training an SNN with multiple time steps. Then, in each phase aimed at reducing latency, the network trained in the previous phase is used as the initialization for subsequent training, gradually reducing the number of time steps. \cite{tang2024onespike}, based on ANN-to-SNN conversion, uses a parallel spike-generation method to convert a multi-time-step SNN into a single time-step SNN while maintaining high accuracy. The work in \cite{suetake2023s3nn} introduces a Single-Step Neural model based on the LIF neuron model, which simplifies multi-time-step SNNs into a single time step, thereby significantly reducing computational costs and latency. The work in \cite{putra2023topspark} analyzes the impact of different time steps on accuracy and uses parameter enhancements and trade-off strategies to optimize the number of time steps. \cite{xu2024stcsnn} proposes a new SNN architecture that introduces spatio-temporal conversion blocks for encoding information into spike sequences and decoding it back to real-valued information, maintaining the low power consumption of the SNN. \cite{xu2023ultra} presents a spatio-temporal compression method that aggregates individual events into a small number of time steps of synaptic current and introduces a Synaptic Convolution Block to balance drastic changes between adjacent time steps, thereby preserving the accuracy of the SNN at a high compression ratio. \cite{pei2023albsnn} introduces a global average pooling layer to replace traditional fully connected layers and proposes an Accuracy Loss Estimator that dynamically selects network layers for binarization to balance the trade-offs between full-precision weights and binarized weights, significantly reducing training and inference time while maintaining high classification accuracy.

However, these methods often involve retraining processes using backpropagation through time in SNNs, or lack neuron models suitable for single-time-step SNNs. Additionally, they introduce complex gradient calculations to compensate for the loss of information in single-time-step SNNs compared to multi-time-step SNNs.
\subsection{Bayesian Optimization Framework}
Current methods for selecting the best hyperparameter set include grid search, random search \cite{bergstra2012random}, and Bayesian optimization \cite{movckus1975bayesian}. Grid search exhaustively evaluates all possible parameter combinations within a predefined parameter grid. While it guarantees finding the optimal solution within the given range, its computational cost grows exponentially with the number of dimensions, making it inefficient. Random search, on the other hand, evaluates the parameter space by randomly sampling it, thereby avoiding the dimensionality curse of grid search. It often finds a better solution with a higher probability under the same computational budget, especially in cases where the importance of parameters varies greatly. However, compared to other search methods, it still requires more time and computational resources.

The core of Bayesian optimization lies in constructing a Gaussian Process (GP) surrogate model to approximate the objective function. It then uses acquisition functions (such as Expected Improvement (EI) and Upper Confidence Bound (UCB)) to intelligently select the next candidate point to evaluate, thereby finding the global optimum with as few evaluations as possible. The Gaussian process is defined as:
\begin{equation}\label{eq:4}
\text{f}(x) \sim \text{GP}(m(x), k(x, x'))
\end{equation}
In modeling an unknown objective function $f(x)$ using Gaussian processes, $m(x)$ represents the mean function (often set as a constant), and $k(x, x')$ is the covariance kernel function, which measures the similarity between two input points $x$ and $x'$. During the optimization process, Bayesian optimization employs an acquisition function (such as Expected Improvement) to balance exploration and exploitation.
\begin{equation}\label{eq:5}
{{\alpha }_{EI}}(x)=E[\max (\text{f}(x)-\text{f}(x^{*}),0)]
\end{equation}
where $\text{f}(x^{*})$ is the current observed optimal function value. The formula calculates the expected improvement $\text{f}(x)-\text{f}(x^{*})$ that a new evaluation point x might bring. This method efficiently performs high-dimensional parameter search in SNN time step optimization by iteratively updating the surrogate model and maximizing the acquisition function, significantly reducing computational costs compared to grid search and random search. This paper employs Bayesian optimization to search for the time step and has conducted further optimization.

%针对与降低网络延时的相关工作
%其中以第三人称的方式把我们的之前的工作也介绍一下
\section{Methodology}
\subsection{Self-Dropping Neuron Model Design}
The LIF neuron is currently the most commonly used type of neuron. When the membrane potential rises to the threshold, it triggers an action potential, thereby generating a spike. In the absence of input, the membrane potential decays exponentially to the resting potential. As the network structure and the number of preceding neurons increase, the membrane potential of subsequent neurons rapidly accumulates. Due to the simple spike-triggering mechanism of conventional neurons, they tend to generate redundant spike sequences. This can cause the neuron to process excessive information or even update to an incorrect state.

\begin{figure}[htbp]
    \centering
    \includegraphics[width=0.9\columnwidth]{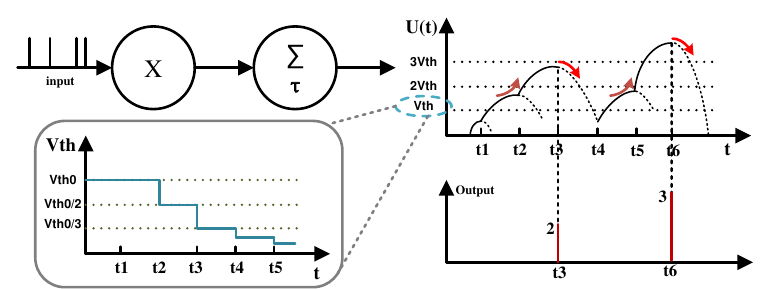}
    \caption{Schematic diagram of spike generation in the Self-Dropping Neuron model}
    \label{fig:SD_neuron_model}
\end{figure}

Fig. 1 illustrates the Self-Dropping (SD) neuron model we proposed for controlling and transmitting information. The specific mechanism is as follows:

\begin{enumerate}
    \item \textbf{Receiving Phase:} The neuron receives input from the upper-layer neurons or the original spike sequence.
    
    \item \textbf{Accumulation Phase:} The neuron calculates the summation of its membrane potential and the input spike sequence to update the current membrane potential.
    
    \item \textbf{Activation Phase:} Based on a predefined initial activation threshold, the membrane potential exceeds the threshold and continues to grow until it begins to decay. When this happens, the neuron triggers a spike. The amplitude of the spike is determined by the ratio of the membrane potential to the threshold. Additionally, based on the complexity of different datasets, a maximum firing limit is set for the spike.
    
    \item \textbf{Backpropagation:} We employ a single-time-step surrogate gradient method, where gradients are calculated only on the spikes in the final time step of the forward propagation. The time-domain gradient computation is not considered to simplify the backpropagation.
    %\hl{thus saving memory and reducing computational complexity. Add Experimental results for it.}
\end{enumerate}

Furthermore, within a single simulation time step, the neuron's membrane potential is recorded. The spike is triggered only when the following conditions are simultaneously met: the membrane potential exceeds the threshold and is lower than the membrane potential of the previous time step. In other cases, no response is given. The ratio of membrane potential to threshold is calculated by dividing the membrane potential by the threshold and rounding down to the nearest integer. This value is then compared with the manually set maximum firing limit, and the smaller value is selected as the final firing result. The update formula used to calculate the membrane potential during the accumulation phase is as follows:
\begin{equation}\label{eq:6}
u^{(t,n+1)} = \gamma u^{(t-1,n+1)} - V_{th} \cdot o^{(t-1,n+1)} + x^{(t,n)}
\end{equation}
\begin{equation}\label{eq:7}
o^{(t,n+1)} = 
\begin{cases}
\left\lfloor \frac{u^{(t,n+1)}}{V_{th}} \right\rfloor & V_{th}< u^{(t,n+1)} < u^{(t-1,n+1)} \\
0 & \text{otherwise}
\end{cases}
\end{equation}
where $V_{th}$ is the threshold, $u^{(t,n+1)}$ is the membrane potential of the n+1 layer at time t, $o$ represents the output, and $\gamma$ is the time constant.
To prevent the occurrence of dead neurons due to small input values not reaching the firing condition, the activation threshold is gradually lowered as the time step increases, as follows:
\begin{equation}\label{eq:8}
V_{th} = \frac{V_{th0}}{t}
\end{equation}
For the loss function, we choose to use Mean Squared Error (MSE), which is the most commonly used error function in regression. It is the average of the squared differences between the predicted value $\text f(x)$ and the target value $y$, and is given by the following formula:
\begin{equation}\label{eq:9}
\text{MSE} = \frac{1}{n} \sum_{i=1}^{n} (f(x_i) - y_i)^2
\end{equation}
We employed the surrogate gradient method to compute the gradients of the spike signals. Specifically, in the spiking layer, we compute a gradient mask to approximate gradient propagation. For each element of the membrane potential $u$, we calculate whether the absolute difference between it and the threshold $Vth$ is less than a predetermined value (which we set to 0.5 by default). If this condition is met, the gradient at that position is retained; otherwise, the gradient is suppressed. The surrogate gradient function is defined as follows:
\begin{equation}\label{eq:10}
h(u) = \frac{1}{a} \, \text{sign} \left( \left| u - V_{th} \right| < \frac{a}{2} \right)
\end{equation}
where $u$ is the membrane potential, sign() is the sign function, $V_{th}$ is the threshold, and $a$ is a parameter that determines the sharpness of the curve, which is set to 1 in this paper.

\subsection{Bayesian Optimization Method for Time Step Selection}
The overall process and setup of the proposed Bayesian method for time step selection are explained as follows. First, we describe the Bayesian configuration. The input \( T(t_1, t_2, \ldots) \) represents the configuration of the spiking layers in the network. The objective function \( f(X) \) that we attempt to maximize is the accuracy of spiking neural networks. 
%We employed the \text{gp\_hedge} function to dynamically select different acquisition functions (Lower Confidence Bound (LCB), Expected Improvement (EI), and Probability of Improvement (PI)) based on the current optimization progress and the model's uncertainty. These acquisition functions assess the potential value of each candidate point to balance exploring unknown areas and exploiting known information, thereby determining the next evaluation point to maximize the likelihood of optimizing the objective function.}\hl{Introduce the principles rather than what libraries and functions will be used}
We dynamically select different acquisition functions (Lower Confidence Bound (LCB), Expected Improvement (EI), and Probability of Improvement (PI)) based on the current optimization progress and the model's uncertainty. These acquisition functions assess the potential value of each candidate point to balance exploring unknown areas and exploiting known information, thereby determining the next evaluation point to maximize the likelihood of optimizing the objective function. The overall process of Bayesian optimization is illustrated below:

\begin{enumerate}
    \item \textbf{Select Initial Points:} Randomly select some initial points in the search space to evaluate the objective function. These initial points provide the initial data for the surrogate model.
    \item \textbf{Construct Surrogate Model:} Using these initial points and their objective function values to construct a Gaussian process as a surrogate model for the objective function.
    \item \textbf{Iterative Optimization:} The optimization is performed iteratively to improve the estimate of the objective function and find the global minimum.    
\end{enumerate}

\subsection{Three-Stage Time Step Bayesian Search Method}
Training a neural network typically requires multiple epochs, and the accuracy generally improves progressively as the number of training epochs increases. Previous studies have shown that the information in the SNN time step exhibits different distributions across different training epochs \cite{kim2023exploring}. Therefore, by identifying the time steps that contain the most information in each epoch, we can reduce time redundancy while improving accuracy. Based on this idea, we propose the Three-Stage Time Step Bayesian Search Method, which aims to search for the optimal combination of time steps across different training epochs. 
\begin{figure}[htbp]
    \centering
    \includegraphics[width=0.4\textwidth, clip]{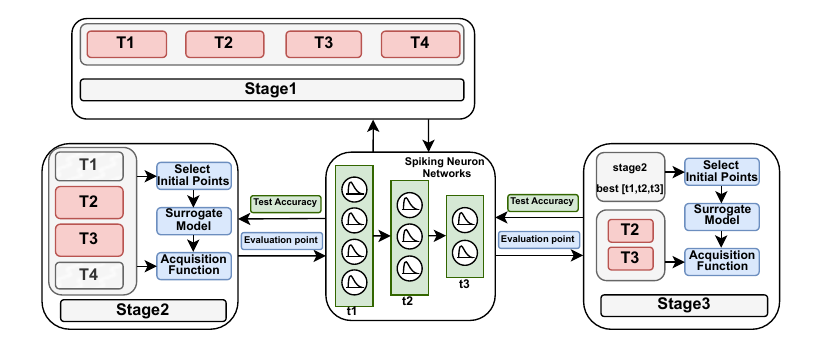}
    \caption{Three-Stage Time Step Bayesian Optimization Framework Diagram}
    \label{fig:bayesian}
\end{figure}
An overview of the method is shown in Fig. \ref{fig:bayesian}, with a detailed description of each stage as follows:
\begin{enumerate}
    \item \textbf{Stage 1: Time Step Search — Global Shared Step} \\
   At this stage, we consider all time steps within the predefined search space, with all spiking layers employing the same n\textsuperscript{th} step size. Using Bayesian optimization, each time step is evaluated, and time steps with low performance are excluded. This process aims to initially filter out promising time steps, narrowing the search space and providing a foundation for optimization in the subsequent stages. After this stage, we obtain the filtered search space for the final Bayesian optimization of time steps.
    
    \item \textbf{Stage 2: Time Step Search — Initial Target Training Epoch} \\
   %\textcolor{blue}{At this stage, we further optimize the selection of time steps between different spiking layers in the SNN, within the search space narrowed down in the first stage, by employing Bayesian optimization. In this stage, we set a smaller number of target training epochs. Through the iterative process of Bayesian optimization, we progressively explore the optimal combination of time steps. In each iteration, Bayesian optimization selects the next evaluation point based on the current surrogate model, evaluates the objective function (test accuracy), and updates the surrogate model. The goal of this stage is to identify a combination of time steps that can quickly enhance model performance within a short period, thereby providing support for subsequent fine-tuning.}
   At this stage, we further optimize the selection of time steps between different spiking layers in the SNN, after the search space is scaled down in the first stage, by employing Bayesian optimization. In this stage, we set a smaller number of target training epochs. Through the iterative process of Bayesian optimization, we progressively explore the optimal combination of time steps. In each iteration, Bayesian optimization selects the next evaluation point based on the current surrogate model, evaluates the objective function (test accuracy), and updates the surrogate model. The goal of this stage is to identify a combination of time steps that can quickly enhance model performance within a short period, thereby providing support for subsequent fine-tuning.
    
    \item \textbf{Stage 3: Time Step Search — Optimization of Recognition Accuracy in the Final Target Epoch} \\
   %\textcolor{blue}{ At this stage, the training epoch is set to the final target epoch, with the goal of optimizing recognition accuracy throughout the entire training process. This optimization is based on Stage 2 and utilizes Bayesian optimization to further refine the selection of time steps between different spiking layers in the SNN within the search space obtained in Stage 1. The optimization process in this stage is the same as in Stage 2, but the goal is to further enhance the model's performance.}
   At this stage, the training epoch is set to the final target epoch to optimize recognition accuracy throughout the entire training process. This optimization is based on Stage 2 and utilizes Bayesian optimization to further refine the selection of time steps between different spiking layers in the SNN within the search space obtained in Stage 1. The optimization process in this stage is the same as in Stage 2, but the goal is to enhance the model's performance further.
\end{enumerate}
%\hl{combine with Fig.2 explain three steps.}

Through the proposed Three-Stage Time Step Bayesian Search Method, we can dynamically adjust the time steps during different training epochs, thereby reducing time redundancy while improving accuracy. 

\subsection{Overall Framework}
This section proposes an overall framework that integrates the Self-Dropping (SD) neuron model with Bayesian optimization methods to enhance the performance and efficiency of Spiking Neural Networks (SNNs). A spiking layer equipped with SD neurons is employed before the convolutional layers in the network to ensure that discrete values are used for convolutional calculations. For the \( l \)-th spiking layer, the input circulates for \( t_l \) time steps, and the information from the \( t_l \)-th time step is subsequently propagated backward. The time step combination for all spiking layers is denoted as \( T = \{t_1, t_2, \ldots, t_l, \ldots, t_n\} \), which is determined by the Bayesian optimization method. The membrane potential information and the threshold at the current time step are saved for backpropagation. As shown in Eq.~(\ref{eq:10}).

\begin{figure}[htbp]
    \centering
    \includegraphics[width=0.4\textwidth, clip]{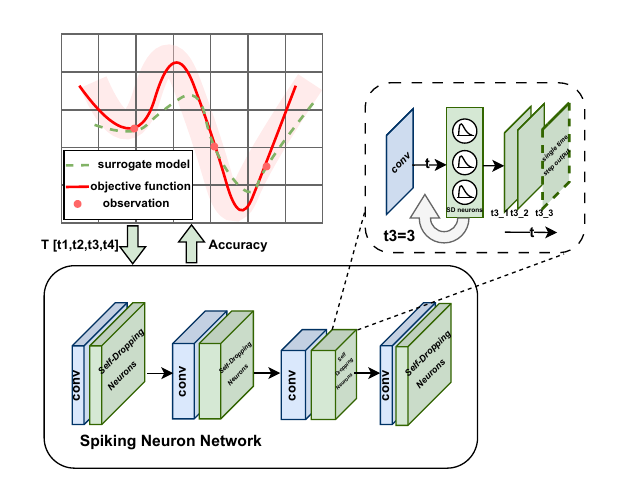}
    \caption{Schematic diagram of the overall network framework}
    \label{fig:overall}
\end{figure}

Following the spiking layer, max pooling is utilized to reduce the image size. Max pooling not only preserves the spiking values but also enhances the fitting capability of SNNs in temporal tasks \cite{Fang_2021_ICCV}. The time steps for all spiking layers in the network are dynamically adjusted through a three-stage Bayesian optimization method.
%\textcolor{blue}{Firstly, all spiking layers uniformly use the same time step, that is, \( T = \{t, \ldots, t, \ldots, t\} \), t is taken from the initial search space.} 
Firstly, all spiking layers uniformly use the same time step, that is, \( T = \{t, \ldots, t, \ldots, t\} \), where $t$ is taken from the initial search space.
%\hl{What does u and v mean? And This letter is easily confused with membrane voltage.}
Each time step is evaluated using the Bayesian optimization method to eliminate poorly performing time steps and preliminarily select promising ones. In the second stage, the time steps are further optimized within a shorter training epoch. In the third stage, the recognition accuracy is optimized within the final target epoch. Through this framework, we can dynamically adjust the time steps across different training epochs to enhance the model's training efficiency and final performance. Additionally, the SD neuron model helps to avoid redundant spiking sequences and erroneous neuron states.

%自回落模型介绍
%引入贝叶斯优化实现时间参数搜索
%整体架构介绍
%采用总分还是分总的结构你自己决定

%一段话总述，可参考：
\subsection{Experimental Setup}
All experiments reported here are conducted on an NVIDIA Tesla V100S GPU. The proposed SDSNN is implemented using the PyTorch framework.
We evaluated the performance of SNNs with SD neurons and Bayesian search for time steps on traditional datasets, including Fashion-MNIST\cite{fashion}, CIFAR-10, and CIFAR-100\cite{cifar10}.
These datasets are widely used standard image classification datasets that can effectively validate the performance of the proposed models.
The network structures for the SNNs on different datasets are shown in Table~\ref{tab:network_structure}.
%\hl{add the simulation environment}
\begin{table}[htbp]
\centering
\begin{threeparttable} 
    \begin{tabular}{@{}lp{5cm}@{}} % 第二列设置为固定宽度，允许自动换行
    \toprule
    \textbf{Dataset} & \textbf{Network Structure} \\ \midrule
    Fashion-MNIST & Input-64C-SD-MP-256C-SD-MP-512C-SD-10C-SD-Voting-10 \\
    CIFAR-10 & Input-128C-SD-256C-SD-MP-512C-SD-MP-1024C-SD-512C-SD-100C-SD-Voting-10 \\
    CIFAR-100 & Input-128C-SD-256C-SD-MP2-512C-SD-MP2-1024C-SD-512C-SD-MP2-100C-SD-Voting-100 \\\bottomrule
    \end{tabular}
\caption{Network structures for different datasets.}
\label{tab:network_structure}
 \begin{tablenotes}    %这行要添加， 从这开始
        \footnotesize               %这行要添加
        \item[1] 64C means a convolutional operation with 64 output channels. SD refers to the proposed SD spiking layer. MP is a max pooling layer with 2x2 filters. Voting refers to the voting layer.
        %\hl{64C3 means XXX, SD refers to , MP is ,Voting refers to }
        %\item[2] The quick brown fox jumps over the lazy dog.        %这行要添加
      \end{tablenotes}            %这行要添加
\end{threeparttable} 

\end{table}

The main structure of the neural network consists of convolutional layers, down-sampling layers, and Batch Normalization layers. The first convolutional layer serves as the encoding layer, receiving raw data input and performing encoding. Down-sampling layers follow the convolutional layers to reduce the computational load. All Conv2D layers are set with a kernel size of 3, a stride of 1, and padding of 1. The pooling layers use max pooling, with all pooling layers set to a kernel size of 2 and a stride of 2. Finally, the voting layers are implemented using average pooling. We use the Adam optimizer and a cosine annealing learning rate scheduler to accelerate the training process. Other hyperparameters of the network are set as shown in Table~\ref{tab:hyperparameters}.
\begin{table}[htbp]
\centering
\begin{threeparttable} 
\begin{tabular}{p{4cm}p{3cm}}
\toprule
\textbf{Hyper-parameters} & \textbf{Value} \\ \midrule
Epoch\_s1 & 20 \\
Epoch\_s2 & 20 \\
Batch size & 8 \\
Learning rate & 0.001 \\
T\_max & 5 \\
$\tau$ & 0.25 \\
$V_{th0}$ & 0.5 \\ \bottomrule
\end{tabular}

\caption{Hyperparameters of the network. }
\label{tab:hyperparameters}
\begin{tablenotes}    %这行要添加， 从这开始
        \footnotesize               %这行要添加
        \item[1] Epoch\textunderscore s1, Epoch\textunderscore s2, and Epoch\textunderscore s3 represent the number of training iterations for the three stages of Bayesian optimization search, respectively. T\textunderscore max  is the upper limit of the set time step, which includes the initial search space for Bayesian optimization. $\tau$ is the neuronal time constant, and $V_{th0}$ is the initial neuronal threshold.
        %\item[2] The quick brown fox jumps over the lazy dog.        %这行要添加
      \end{tablenotes}            %这行要添加
\end{threeparttable} 

\end{table}

\begin{table*}[htbp]
\centering
\small
\begin{threeparttable} 
\begin{tabular}{@{}lcccccccc@{}}
\toprule
\textbf{Dataset} & \textbf{Neuron Model} & \textbf{T\(_{s1}\)} & \textbf{Acc (\%)} & \textbf{T\(_{s2}\)} & \textbf{Acc (\%)} & \textbf{Energy Efficiency (mJ)} & \textbf{Spiking Rate} \\
\midrule
\multirow{3}{*}{\textbf{Fashion-MNIST}} 
 & IF & 1 & 92.88 & 1 & 93.35 & 0.021 & 15.84 \\
 & LIF & 3 & 93.12 & 3 & 93.51 & 0.050 & 14.58 \\
 & SD & [1,2,1,4] & \textbf{93.22} & [1,3,1,1] & \textbf{93.72} & \textbf{0.022$(\downarrow 56\%)$} & 14.86 \\
\midrule
\multirow{3}{*}{\textbf{CIFAR-10}} 
 & IF & 1 & 90.38 & 1 & 91.90 & 0.027 & 12.87 \\
 & LIF & 2 & 90.13 & 2 & 92.01 & 0.056 & 13.56 \\
 & SD & [1,2,2,2,1,1] & \textbf{91.47} & [1,2,2,2,2,1] & \textbf{92.20} & \textbf{0.044$(\downarrow 21\%)$} & 13.66 \\
\midrule
\multirow{3}{*}{\textbf{CIFAR-100}} 
 & IF & 1 & 57.69 & 1 & 64.62 & 0.035 & 12.74 \\
 & LIF & 2 & 64.04 & 2 & \textbf{70.42} & 0.064 & 12.93 \\
 & SD & [1,2,2,2,2,2] & \textbf{67.47} & [1,2,2,1,2,1] & 69.45 & \textbf{0.050$(\downarrow 22\%)$} & 13.52 \\
\bottomrule
\end{tabular}
\caption{Performance of different neuron models on various datasets. For the SD neuron model}
\label{tab:neuron_models}
\begin{tablenotes}    %这行要添加， 从这开始
        \footnotesize               %这行要添加
        \item[1] \( T_{s1} \) and \( T_{s2} \) represent the time step combinations of the spiking layers obtained from Bayesian optimization in the second and third stages, respectively. For IF and LIF, the time step for each spiking layer is a fixed value.
      \end{tablenotes}            %这行要添加
\end{threeparttable} 

\end{table*}

\begin{table*}[t!]
\centering
\small
\begin{threeparttable} 
\begin{tabular}{@{}>{\raggedright\arraybackslash}p{2cm}>{\centering\arraybackslash}p{2cm}>{\centering\arraybackslash}p{4cm}>{\centering\arraybackslash}p{2cm}>{\centering\arraybackslash}p{4cm}@{}}
\toprule
\textbf{Dataset} & \textbf{T\(_{s1}\)} & \textbf{Accuracy (\%)} & \textbf{T\(_{s2}\)} & \textbf{Accuracy (\%)} \\
\midrule
\multirow{2}{*}{\textbf{Fashion-MNIST}} 
 & 2 / 3 / 4 / 5 & 92.42 / 92.10 / 91.98 / 92.56 & 2 / 3 / 4 / 5 & 92.69 / 92.20 / 92.17 / 92.77 \\
 & [1,2,1,4] & \textbf{93.22} & [1,3,1,1] & \textbf{93.72} \\
\midrule
\multirow{2}{*}{\textbf{CIFAR-10}} 
 & 2 / 3 / 4 / 5 & 90.45 / 87.90 / 87.54 / 86.72 & 2 / 3 / 4 / 5 & 84.44 / 90.52 / 90.36 / 90.37 \\
 & [1,2,2,2,1,1] & \textbf{91.47} & [1,2,2,2,2,1] & \textbf{92.20} \\
\midrule
\multirow{2}{*}{\textbf{CIFAR-100}} 
 & 2 / 3 / 4 / 5 & 66.59 / 58.14 / 58.17 / 38.72 & 2 / 3 / 4 / 5 & 69.84 / 62.51* / 63.26 / 41.96* \\
 & [1,2,2,2,2,2] & \textbf{67.47} & [1,2,2,1,2,1] & \textbf{69.45} \\
\bottomrule
\end{tabular}
\caption{Performance of the network using the SD neuron model under different time step methods}
\label{tab:sd_neuron_models}
\begin{tablenotes}    %这行要添加， 从这开始
        \footnotesize               %这行要添加
        \item[1] It is worth noting that the spike firing mechanism of the SD neuron relies on comparing the membrane potential value from the previous time step. There is no previous value available for comparison at the first time step; the IF method is used for spike firing in the first time step of the SD neuron model. Marked with * indicates that the time steps were excluded by the first-stage Bayesian optimization, retaining only promising time steps and eliminating those with significant deviations.
      \end{tablenotes}            %这行要添加
\end{threeparttable} 

\end{table*}

\begin{table*}[t!]
\centering
\small % 可选：缩小字体使表格更紧凑
\begin{tabular}{@{}llllcc@{}} % 使用@{}减少两侧空白
\toprule
\textbf{Dataset} & \textbf{Model} & \textbf{Method} & \textbf{Architecture} & \textbf{Timestep} & \textbf{Accuracy (\%)} \\
\midrule
\multirow{4}{*}{\textbf{Fashion-MNIST}}
 & \cite{cheng2020lisnn} & SNN & 2Conv, 1Linear & 20 & 92.07 \\
 & \cite{NEURIPS2020_8bdb5058} & SNN & 2Conv, 1Linear & 5 & 92.83 \\
 & \cite{pei2023albsnn} & ALBSNN & 6Convs & 1 & 93.10 \\
 & \textbf{This work} &SDSNN & 4Conv & 1 & 93.72 \\
\cmidrule(r){1-6}
\multirow{4}{*}{\textbf{CIFAR-10}}
 & \cite{NEURIPS2020_8bdb5058} & SNN & 5Conv, 2Linear & 5 & 91.41 \\
 & \cite{chen2021bnn} & BinaryNN & ReActNet18 & 1 & 92.08 \\
 & \cite{pei2023albsnn} & ALBSNN & 2Conv & 1 & 90.12 \\
 & \textbf{This work} & SDSNN & 6Conv & 1 & 92.20 \\
\cmidrule(r){1-6}
\multirow{4}{*}{\textbf{CIFAR-100}}
 & \cite{rathi2021diet} & Diet-SNN & VGG16 & 5 & 69.67 \\
 & \cite{chen2021bnn} & BinaryNN & ReActNet18 & 1 & 68.79 \\
 & \cite{pei2023albsnn} & ALBSNN & 3Conv & 1 & 63.54 \\
 & \textbf{This work} & SDSNN & 6Conv & 1 & 69.45 \\
\bottomrule
\end{tabular}
\caption{Compare with exciting work on different datasets.}
\label{tab:comparison}
\end{table*}

\subsection{Performance Analysis}
%计算复杂度、精度、脉冲稀疏度、功耗、延时等方面
In this section, we will explore various enhancements to the neuron models and time-step algorithms. Specifically, we focus on the classical LIF (Leaky Integrate-and-Fire) and IF (Integrate-and-Fire) neuron models, as well as the SD (Self-Dropping) model we proposed. Additionally, we will explore the performance of single-step and multi-step SNNs in different scenarios. Our goal is to comprehensively evaluate the performance of these methods from multiple angles, including accuracy and power consumption.

To systematically evaluate the performance of different neuron models and time-step algorithms, we designed a series of experiments. These experiments cover multiple datasets to ensure the broad applicability of the results. We not only focus on the accuracy of the models but also on power consumption, a key metric that directly affects the feasibility and efficiency of SNNs in practical applications.

Table~\ref{tab:neuron_models} presents a comparison of accuracy between the proposed method and other methods on the FashionMNIST, CIFAR-10, and CIFAR-100 datasets. In all experiments, we employed Bayesian optimization to conduct 100 searches, ensuring thorough exploration of the parameter space and identification of the optimal solution. We have conducted a detailed evaluation of the power consumption of our study using the method proposed in \cite{10.1007/978-3-031-30105-6_48}. 
%\textcolor{blue}{Compared to SNN using LIF, in the stage 3 on the Fashion-MNIST dataset, the SDSNN achieved a 0.21\% increase in accuracy and a 56\% reduction in energy consumption. On the CIFAR-10 dataset, in the stage 2, the SDSNN achieved a 1.34\% increase in accuracy and a 21\% reduction in energy consumption.}
Compared to SNN using LIF, in stage 3 on the Fashion-MNIST dataset, the SDSNN achieved a 0.21\% increase in accuracy and a 56\% reduction in energy consumption. On the CIFAR-10 dataset, in stage 2, the SDSNN achieved a 1.34\% increase in accuracy and a 21\% reduction in energy consumption.
Through experiments on various datasets, it has been found that our method exhibits competitive performance when using  only a single time step and achieves a significant reduction in power consumption.
%\textcolor{blue}{The SD model has a spiking rate similar to that of the LIF model, but unlike the multi-timestep approach of LIF, our single-timestep network involves only the convolution calculation of a single layer of spike information, thereby reducing computational overhead.}
The SD model has a spiking rate similar to that of the LIF model. Still, unlike the multi-timestep LIF, our single-timestep network involves only the convolution calculation of a single layer of spike information, thereby reducing computational overhead.
%\hl{Add comparative analysis of representative data to draw conclusions}

Table~\ref{tab:sd_neuron_models} shows a comparison between traditional fixed time steps (all spiking layers use the same n\textsuperscript{th} time step) and adaptive time step configurations based on Bayesian optimization. 
%\textcolor{blue}{On CIFAR-10, compared to the fixed time-step method, the Bayesian optimization method in Stage 2 increased the optimal result from 90.45\% to 91.47\%, which validates the effectiveness of the proposed adaptive time-step optimization method.} 
On CIFAR-10, compared to the fixed time-step method, the Bayesian optimization method in Stage 2 increased the optimal result from 90.45\% to 91.47\%, which validates the effectiveness of the proposed adaptive time-step optimization method.
Table~\ref{tab:comparison} presents the comparative analysis between our proposed method and existing approaches.
%\textcolor{blue}{On CIFAR-10, the proposed method achieves higher accuracy than multi-timestep SNNs and other single-timestep SNNs by 0.79\%, 0.12\%, and 2.08\%, respectively. On CIFAR-100, the accuracy is slightly lower than that of multi-timestep SNNs.}
On CIFAR-10, the proposed method achieves higher accuracy than multi-timestep SNNs and other single-timestep SNNs by 0.79\%, 0.12\%, and 2.08\%, respectively. On CIFAR-100, the accuracy is slightly lower than that of multi-timestep SNNs with significant energy consumption reduction.
%\hl{Add comparative analysis of representative data to draw conclusions}

%\textcolor{blue}{We measure the peak memory usage during training using PyTorch's memory monitoring APIs \cite{paszke2019pytorch}. As shown in Table 6, the proposed single-timestep SNN only needs to store the membrane potential and threshold of one spiking layer, reducing the theoretical memory complexity from O(T) to O(1). On CIFAR-10, our method consumes only 306.23MB memory, demonstrating significant reduction compared to BPTT.}
We measure the peak memory usage during training using PyTorch's memory monitoring APIs \cite{paszke2019pytorch}. As shown in Table 6, the proposed single-timestep SNN only needs to store the membrane potential and threshold of one spiking layer, reducing the theoretical memory complexity from O(T) to O(1). On CIFAR-10, our method consumes only 0.91G of memory, demonstrating a significant reduction compared to BPTT.

\begin{table*}[t!]
\centering
\small
\begin{threeparttable} 
\begin{tabular}{@{}llllcc@{}}
\toprule
\textbf{Dataset} & \textbf{Model} & \textbf{Method} & \textbf{Architecture} & \textbf{Timestep} & \textbf{Memory (GB)} \\
\midrule
\multirow{3}{*}{\textbf{CIFAR-10}} 
 & \cite{werbos2002backpropagation} & BPTT & ResNet-18 & 6 & 3.00 \\
 & \cite{meng2023towards} & SLTT & ResNet-18 & 6 & 1.09 \\
 & This work & SDSNN & ResNet-18 & 1 & 0.91 \\
\cmidrule(r){1-6}
\multirow{3}{*}{\textbf{CIFAR-100}}
 & \cite{werbos2002backpropagation} & BPTT & ResNet-18 & 6 & 3.00 \\
 & \cite{meng2023towards} & SLTT & ResNet-18 & 6 & 1.12 \\
 & This work & SDSNN & ResNet-18 & 1 & 1.02 \\
\bottomrule
\end{tabular}
\caption{ GPU memory usage comparison}
\label{tab:memory_comparison}
\end{threeparttable}
\end{table*}

\section{Discussions and Conclusion}
In this work, we propose the Self-Dropping (SD) neuron, which adjusts the neuron’s spike emission mechanism. Specifically, we modify the traditional mechanism of triggering a spike when the membrane potential exceeds the threshold. Instead, we propose that a spike is triggered when the membrane potential exceeds the initial threshold and continues to rise until it starts to decay. This approach does not require an additional loss function to guide the learning process; instead, it introduces variable spikes during the training process while limiting the maximum spike threshold. This design enhances the representation capacity of the spikes while balancing both temporal and spatial demands.
Furthermore, we introduce a Three-Stage Time Step Search Method, which utilizes Bayesian optimization to find the optimal time step for each spiking layer. This staged search method accounts for the differences in temporal concentration of information during different training periods, ensuring that spikes with a single time step propagate to subsequent layers. This single-step information transmission in SNNs significantly reduces latency and power consumption, while also enabling faster convergence and achieving better accuracy.


\bibliography{aaai2026}

\end{document}